\pdfoutput=1
\documentclass[11pt]{article}

\usepackage[final]{ACL2023}

\usepackage{times}
\usepackage{latexsym}
\definecolor{c1}{HTML}{ed1941}
\definecolor{c2}{HTML}{2485a6}

\usepackage{threeparttable}
\usepackage{multirow, makecell}
\usepackage{url}
\usepackage{csquotes}
\usepackage{bbding}
\usepackage{lineno}
\usepackage{enumitem}
\usepackage{graphicx}
\usepackage{booktabs}
\usepackage{amssymb}
\usepackage[ruled,vlined,linesnumbered]{algorithm2e}
\SetAlFnt{\scriptsize}
\SetAlCapFnt{\small}
\SetAlCapNameFnt{\small}

\usepackage[T1]{fontenc}

\usepackage[utf8]{inputenc}

\usepackage{microtype}
\usepackage{inconsolata}
\usepackage{comment}

\usepackage{snaptodo}
\snaptodoset{block rise=2em}
\snaptodoset{margin block/.style={font=\tiny, fill=white}}

\newcommand{\squishlist}{
 \begin{list}{$\bullet$}
  { \setlength{\itemsep}{0pt}
     \setlength{\parsep}{3pt}
     \setlength{\topsep}{3pt}
     \setlength{\partopsep}{0pt}
     \setlength{\leftmargin}{1.5em}
     \setlength{\labelwidth}{1em}
     \setlength{\labelsep}{0.5em} } }

\newcommand{\squishend}{
  \end{list}  }

\title{Knowledge Base Completion for Long-Tail Entities}

\author{
Lihu Chen\textsuperscript{\rm 1 \thanks{ \quad Work done during an internship at Max Planck Institute for Informatics}}, 
Simon Razniewski\textsuperscript{\rm 2}, 
Gerhard Weikum\textsuperscript{\rm 2} \\
\textsuperscript{\rm 1} LTCI, Télécom Paris, Institut Polytechnique de Paris, France \\
\textsuperscript{\rm 2} Max Planck Institute for Informatics, Saarbrücken, Germany \\
\texttt{\{lihu.chen\}@telecom-paris.fr}\\ 
\texttt{\{srazniew,weikum\}@mpi-inf.mpg.de}
}

\begin{document}
\maketitle
\begin{abstract}
Despite their impressive scale, 
knowledge bases (KBs), such as Wikidata, still contain significant gaps. Language models (LMs) have been proposed as a source for filling these gaps.
However, prior works have focused on prominent entities with rich coverage by LMs, neglecting
the crucial case of long-tail entities.
In this paper, we present a novel method for LM-based-KB completion that is specifically geared for facts about long-tail entities. The method leverages two different LMs in two stages: for candidate retrieval and for candidate verification and disambiguation. 
To evaluate our method and various baselines, 
we introduce a novel dataset, called MALT, rooted in Wikidata. Our method outperforms all baselines in F1, with major gains especially in recall.
\end{abstract}

\section{Introduction }

{\bf Motivation and Problem.}
Knowledge base completion (KBC) is crucial
to continuously enhance the scope and scale of large
knowledge graphs (KGs). It is often cast into a link prediction
task: infer an O(bject) argument for a given S(ubject)-P(redicate) pair.
However, the task is focused on the KG itself as the only input,
and thus largely bound to predict SPO facts that are also derivable by simple logical rules for inverse predicates, transitive predicates etc.
\cite{DBLP:conf/sigmod/AkramiSZHL20,DBLP:conf/acl/SunVSTY20}. To obtain truly new facts, more recent methods tap into
large language models (LMs) that are learned from huge text collections,
including all Wikipedia articles, news articles and more.
The most promising approaches to this end generate cloze questions
for knowledge acquisition and ask LMs to generate answers \cite{petroni2019language}.
The LM input is often augmented with carefully crafted short prompts 
(e.g., a relevant Wikipedia paragraph) \cite{shin2020autoprompt,jiang2020can,DBLP:conf/naacl/QinE21}.

However, notwithstanding great success for question answering to humans,
the LM-based approach falls short on meeting the high quality requirements
for enriching a KG with crisp SPO facts. Even if most answers are correct,
there is a non-negligible fraction of false or even ``hallucinated'' outputs
by the LM, and large KGs, like Wikidata~\cite{vrandevcic2014wikidata}, cannot tolerate error rates
above 10 percent. Moreover, even correct answers are not properly canonicalized:
they are surface phrases and not unique entities in the KG. 
These problems are further aggravated when the to-be-inferred O arguments
are {\em long-tail} entities, with very few facts in Wikidata.
Here, we call an entity \emph{long-tail} when it has less than 14 triples in Wikidata, because nearly 50\% of the Wikidata entities have fewer than 14 triples.
These are exactly the pain point that calls for KBC. This paper 
addresses this problem.

As an example, consider the late Canadian singer \textit{Lhasa de Sela}.
Wikidata solely covers basic biographic facts and selected awards,
nothing about her music. However, text sources such as her Wikipedia article
or other web pages provide expressive statements about her albums, songs,
collaborations etc. For example, we would like to spot the facts that
$\langle$\textit{Lhasa de Sela, collaboratedWith, Bratsch}$\rangle$ and
$\langle$\textit{Lhasa de Sela, performedSong, Anyone and Everyone}$\rangle$.
Note that capturing these as SPO facts faces the challenge of
having to capture and disambiguate multi-word names (\textit{``Lhasa de Sela''}) and
common-noun phrases (\textit{``anyone and everyone''}).
When trying to extract such statements via cloze questions or more refined prompts
to LMs such as GPT-3~\cite{brown2020language} or chatGPT, the outputs would often be \textit{``Lhasa''}, which is highly ambiguous,
or \textit{``everyone''}, which is incomplete and impossible to interpret.

\begin{figure*}[t]
	\centering
	\includegraphics[width=1.0\textwidth]{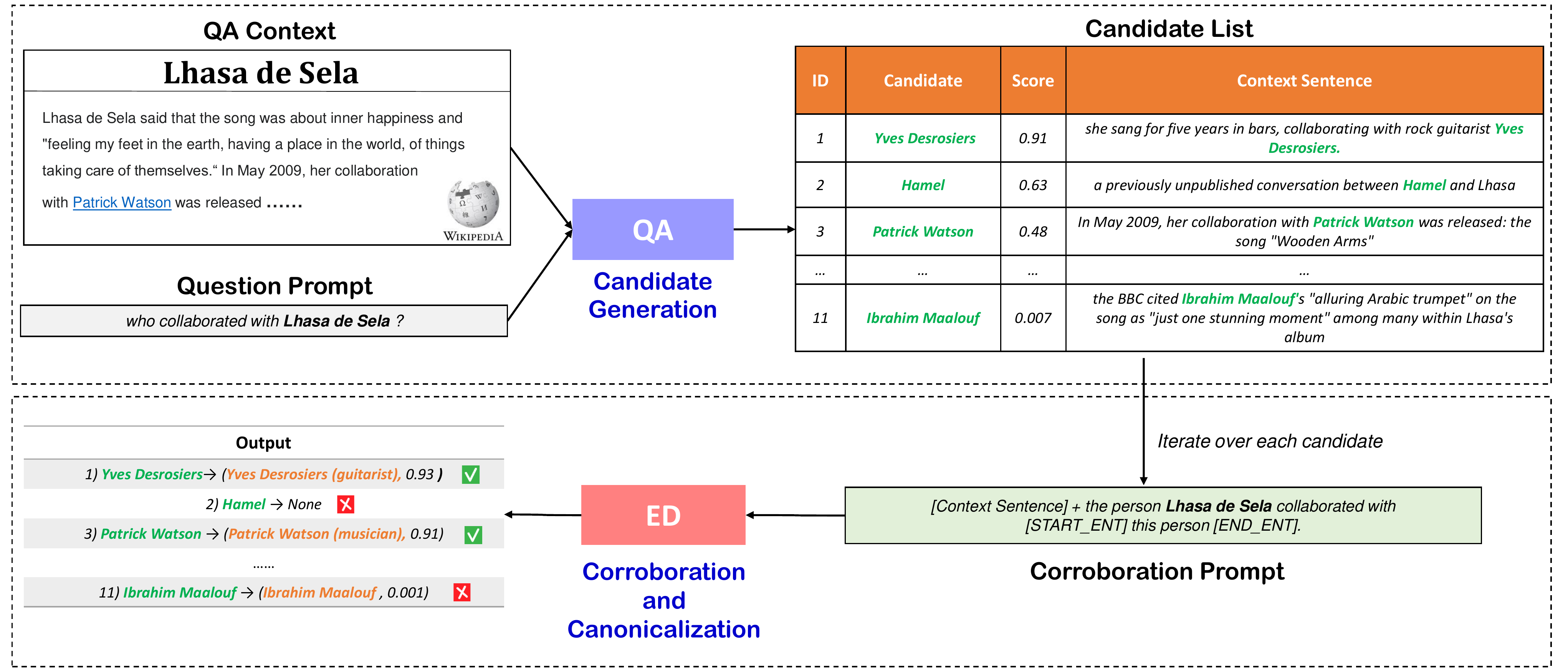}
	\caption{The framework of our two-stage KBC method.}
	\label{fig:framework}
\end{figure*}

\noindent
{\bf Approach and Contribution.}
This paper devises a novel method for knowledge base completion (KBC),
specifically geared to cope with long-tail entities.
Although we will present experimental comparisons to prior works on
relation extraction from text, we believe that ours is among the first works
to successfully cope with the challenge of noise and ambiguity in the long tail.

Our method leverages Transformer-based language models in a new way.
Most notably, we employ two different LMs in a two-stage pipeline, as shown in Figure~\ref{fig:framework}.
The first stage generates candidate answers to input prompts and gives
cues to retrieve informative sentences from Wikipedia and other sources.
The second stage validates (or falsifies) the candidates and disambiguates
the retained answer strings onto entities in the underlying KG
(e.g., mapping \textit{``Lhasa''} to {\small\tt Lhasa de Sela} , and
``Bratsch'' to {\small\tt Bratsch (band)}).

The novel contributions of this work are the following:
\squishlist
\item the first 
KBC method that leverages LMs 
to cope with long-tail entities;
\item a new dataset, called MALT, to benchmark methods with long-tail entities;
\item experimental comparisons with baselines, using the MALT data.
\squishend
Our code and data are available on both GitHub\footnote{\url{https://github.com/tigerchen52/long_tail_kbc}} and mpi-inf.mpg.de\footnote{\url{https://www.mpi-inf.mpg.de/departments/databases-and-information-systems/research/knowledge-base-recall/lm4kbc}}.


\section{Related Work}

{\bf Knowledge Base Completion.}
This task, KBC for short, has mostly been tackled as a form of link prediction: given a head entity S and a relation P, predict the respective tail entity O,
using the KG as sole input.
A rich suite of methods have been developed for this
task, typically based on latent embeddings computed via
matrix or tensor factorization, neural auto-encoders, graph neural networks, and more (see, e.g., surveys \cite{DBLP:journals/access/ChenWZCZD20,DBLP:journals/tnn/JiPCMY22} and original references given there).
However, the premise of inferring missing facts from the KG itself is a fundamental limitation. Indeed, several studies have found that many facts predicted via the above KBC techniques are fairly obvious and could also be derived by simple rules for transitivity, inverse relations etc. \cite{DBLP:conf/sigmod/AkramiSZHL20,DBLP:conf/acl/SunVSTY20}.

\noindent
{\bf Language Models as Knowledge Bases.}
The LAMA project \cite{petroni2019language} posed
the hypothesis that probing LMs with cloze questions is a powerful way of extracting structured facts from the latently represented corpus on which the LM was trained. A suite of follow-up works pursued this theme
further and devised improvements and extensions
(e.g., \cite{heinzerling2021language,jiang2020x,kassner2020negated,roberts2020much,shin2020autoprompt,DBLP:conf/naacl/ZhongFC21}).
This gave rise to the notion of ``prompt engineering'' for all kinds of NLP tasks \cite{DBLP:journals/corr/abs-2107-13586}.
In parallel, other works studied biases and limitations of the LM-as-KB paradigm (e.g., \cite{cao2021knowledgeable,DBLP:journals/tacl/ElazarKRRHSG21,razniewski2021language,jiang2020can}). In this work, we investigate the feasibility of leveraging LMs to complete real-world KBs, and mainly focus on long-tail facts.

\begin {table*}[t] 
\centering\scriptsize
\setlength{\tabcolsep}{2.2mm}{
	\begin{threeparttable} 
		\begin{tabular}{ccccccc}  
			\toprule  
			\textbf{Subject Type}&\textbf{Relation}&\textbf{Wikidata ID}&\textbf{Triples}&\textbf{multi-token (\%)}&\textbf{ambiguous (\%)}&\textbf{long-tail (\%)}\cr
			\midrule
			\texttt{Business}&\texttt{founded by}&P112&5720&97.3&21.1&91.2 \cr
			\midrule
			\multirow{2}{*}{\texttt{MusicComposition}} &\texttt{performer}&P175&1876&91.1&62.0&47.3  \cr
			&\texttt{composer}&P86&3016&98.2&59.8&88.5  \cr
			\midrule
			\multirow{5}{*}{\texttt{Human}} &\texttt{place of birth}&P19&13416&23.6&81.6&99.3 \cr
			&\texttt{place of death}&P20&7247&25.9&84.8&99.6  \cr
			&\texttt{employer}&P108&3503&96.5&37.4&81.4 \cr
			&\texttt{educated at}&P69&13386&99.6&38.7&72.2 \cr
			&\texttt{residence}&P551&886&32.1&87.1&96.4 \cr
			\midrule
			Micro-Avg&-&-&-&65.3&58.6&87.0\cr
			\bottomrule
		\end{tabular}
		\caption{Statistics for MALT dataset.
		} 	\label{tab:statistics}
	\end{threeparttable}
}
\end{table*}

\begin{table}[tbp] 
   \small
    \setlength{\tabcolsep}{2.3mm}{
	\begin{threeparttable} 
		\begin{tabular}{ccc}  
			\toprule    	
			Dataset&SPO triples&Long-tail fraction\cr
			\midrule
                DocRED~\shortcite{yao2019docred}&63K&32.0 \%\cr
                LAMA-TREx~\shortcite{petroni2019language}&34K&39.6 \%\cr
                X-FACTR~\shortcite{jiang2020x}&46K&49.6 \%\cr
                MALT (Ours)&49K&87.0 \%\cr
			\bottomrule
			\end{tabular}
                
			\caption{Estimated fractions of long-tail S entities across different datasets, where long-tail means at most 13 triples in Wikidata. The  estimations are based on 200 samples across 8 relations.}
   \label{tab:dataset_long_tail}
		\end{threeparttable}
	}
\end{table}
\section{Two-Stage KBC Method}

We propose
an unsupervised method for KBC that taps into LMs as latent source for facts that cannot be inferred from the KG itself. 
Our method operates in two stages:
\squishlist
\item[1.] For a given S-P pair, generate candidate facts $\langle$S,P,``O''$\rangle$ where ``O'' is an entity name and possibly a multi-word phrase.
\item[2.] Corroborate the candidates, retaining the ones with high confidence of being correct, and disambiguate the ``O'' argument into a KG entity.
\squishend

\paragraph{Candidate Generation.}
We devise a generic prompt template for cloze questions, in order to infer an ``O'' answer for a given S-P pair. This merely requires a simple verbalizer for the relation P:

\quad\vspace*{0.1cm}
\noindent ``$\langle$S-type$\rangle$ S 
$\langle$P-verb$\rangle$ 
which $\langle$O-type$\rangle$?''
\vspace*{0.1cm}

\noindent (e.g., ``the song $\langle$S$\rangle$ is performed by which person?'' for the predicate {\tt performer}).
The S-type and O-type are easily available by the predicate type-signature from the KG schema.
As additional context we feed a Wikipedia sentence from the S entity's article into the LM. 
This is repeated for all sentences in the respective Wikipedia article. 
Specifically, we employ the SpanBERT language model \cite{joshi2020spanbert}, which is fine-tuned on on the SQuAD 2.0~\cite{rajpurkar2018know}~\footnote{\url{https://huggingface.co/mrm8488/spanbert-large-finetuned-squadv2}}.
Note that all of this is completely unsupervised:
there is no need for any fine-tuning of the LM,
and there is no prompt engineering.

\paragraph{Candidate Corroboration and Canonicalization.}
The first stage yields a scored list
of candidates in the form of pairs (``O'', $s$) with an entity name and a Wikipedia sentence $s$.
In the corroboration stage, the candidates are fed into a second LM for re-ranking and pruning false positives. Specifically, we employ the generative entity disambiguation model GENRE~\cite{de2020autoregressive}, which in turn is based on BART \cite{lewis2020bart} and fine-tuned on BLINK~\cite{wu2020scalable} and AIDA~\cite{hoffart2011robust}.
We construct the input by the template:

\vspace*{0.1cm}
\noindent ``$\langle$S-type$\rangle$ S 
$\langle$P-verb$\rangle$ 
[ENT] this $\langle$O-type$\rangle$ [ENT]''
\vspace*{0.1cm}

\noindent (e.g., ``the song Anyone and Everyone is performed by [ENT] this person [ENT]''),
contextualized with the sentence $s$. 
GENRE generates a list of answer entities
{\small\tt E},
taken from an underlying KG, like Wikidata,
that is, no longer just surface names.
If the candidate name ``O'' approximately matches
a generated {\small\tt E} (considering alias names
provided by the KG), then the entire fact, now properly canonicalized, is kept.
Since we may still retain multiple facts for the same S-P input and cannot perfectly prevent false positives, the inferred facts are scored by
an average of the scores from stage 1 and stage 2.


\begin{table*}[ht] 
	\centering 
	\small
	\setlength{\tabcolsep}{2.2mm}{
		\begin{threeparttable} 
			\begin{tabular}{p{2.5cm} p{0.6cm} p{5.2cm} p{5.2cm}}  
				\toprule
				Relation&ID&Candidate Generation& Corroboration and Canonicalization\cr
				\midrule
                    founded by  &P112&the business \texttt{[x]} is founded by which person?&the business \texttt{[x]} is founded by \texttt{[ENT]} this person \texttt{[ENT]}\cr
                    \midrule
				performer  &P175&the song \texttt{[x]} is performed by which person?&the song \texttt{[x]} is performed by \texttt{[ENT]} this person \texttt{[ENT]}\cr
                \midrule
				composer  &P86&the song \texttt{[x]} is composed by which person?&the song \texttt{[x]} is composed by \texttt{[ENT]} this person \texttt{[ENT]}\cr
                \midrule
				place of birth  &P19&the person \texttt{[x]} was born in which place?&the person \texttt{[x]} was born in \texttt{[ENT]} this place \texttt{[ENT]}\cr
                \midrule
				place of death  &P20&the person \texttt{[x]} died in which place?&the person \texttt{[x]} died in \texttt{[ENT]} this place \texttt{[ENT]}\cr
                \midrule
				employer  &P108&the person \texttt{[x]} worked in which place?&the person \texttt{[x]} worked in \texttt{[ENT]} this place \texttt{[ENT]}\cr
                \midrule
				educated at  &P69&the person \texttt{[x]} graduated from which place?&the person \texttt{[x]} graduated from \texttt{[ENT]} this place \texttt{[ENT]}\cr
                \midrule
				residence  &P551&the person \texttt{[x]} lived in which place?&the person \texttt{[x]} lived in \texttt{[ENT]} this place \texttt{[ENT]}\cr
				\bottomrule
			\end{tabular}
			\caption{Prompts for relations in MALT. \texttt{[x]} is a placeholder for the subject entity and \texttt{[ENT]} is a special token for the mention.}	\label{tab:malt_template}
		\end{threeparttable}
	}
\end{table*}

\begin{table*}[t] 
	\centering\scriptsize 
	\setlength{\tabcolsep}{1.4mm}{
		\begin{threeparttable} 
			\begin{tabular}{cc|ccc|ccc|ccc|ccc|ccc}  
				\toprule  
			\textbf{Relation}&\textbf{ID}&\multicolumn{3}{c|}{\textbf{NER + RC (CNN)  }}&\multicolumn{3}{c|}{\textbf{REBEL}}&\multicolumn{3}{c|}{\textbf{KnowGL}}&\multicolumn{3}{c|}{\textbf{GenIE}}&\multicolumn{3}{c}{\textbf{Ours}}\cr
				&&P&R&F1&P&R&F1&P&R&F1&P&R&F1&P&R&F1\cr
				\midrule
				\texttt{founded by}& P112&  13.5&21.2&16.5&42.8&27.3&33.3&0.0&0.0&0.0&59.1&7.9&13.9&57.0&44.5&50.0\cr
				\midrule
				\texttt{performer} &P175&  5.2&10.1&6.9&25.3&28.1&26.6&0.0&0.0&0.0&47.3&19.1&27.2&42.7&15.6&22.9\cr
				\texttt{composer} &P86&  17.3&20.5&18.8&37.9&27.7&32.0&37.6&25.7&30.6&70.0&16.6&26.8&67.3&65.6&66.4\cr
				\midrule
				\texttt{place of birth} &P19  &4.7&4.7&4.7&49.3&20.5&28.9&49.4&23.4&31.7&64.1&9.2&16.1&47.9&61.4&53.8\cr
				\texttt{place of death}& P20  &12.5&4.7&6.8&52.6&11.8&19.2&66.6&9.4&16.5&47.5&3.0&5.6&46.6&48.2&47.4\cr
				\texttt{employer} &P108  &8.7&4.9&6.3&50.0&4.9&8.8&0.0&0.0&0.0&54.0&0.1&0.2&30.0&29.3&29.6\cr
				\texttt{educated at} &P69  &8.9&8.4&7.7&15.4&1.1&2.1&22.2&1.1&2.2&46.7&0.1&0.2&42.9&39.5&41.2\cr
				\texttt{residence} &P551  &0.0&0.0&0.0&33.3&8.3&13.3&33.3&8.3&13.3&44.4&0.2&0.4&19.2&41.7&26.3\cr
				\midrule
				Micro-Avg &-&26.7&13.7&13.7&38.3&16.2&20.6&26.2&8.5&11.8&52.2&6.9&11.2&44.2&43.2&42.2\cr
				\bottomrule
			\end{tabular}
\caption{Performance comparison on MALT data.}
\label{tab:overall_performance}
		\end{threeparttable}
	}
\end{table*}

\section{MALT: New Dataset for Benchmarking}

Benchmarks for KBC and LM-as-KB cover facts for all kinds of entities, but tend to focus on prominent ones with frequent mentions. Likewise, benchmarks for 
relation extraction (RE) from text, most notably TACRED~\cite{zhang2017position}, DocRED \cite{yao2019docred} and LAMA~\cite{petroni2019language}
do not reflect the difficulty of coping with long-tail entities and the amplified issue of surface-name ambiguity (see Table~\ref{tab:dataset_long_tail}.
Therefore, we developed a new dataset with
emphasis on the long-tail challenge,
called MALT (for 
``\textbf{M}ulti-token, \textbf{A}mbiguous, \textbf{L}ong-\textbf{T}ailed facts'').

To construct the dataset, we focus on 
three types of entities:
\texttt{Business}, 
\texttt{MusicComposition} and \texttt{Human}, richly covered in Wikidata and often involving long-tail entities. 
We randomly select subjects from the respective relations in Wikidata, and keep all objects for them. 
We select 
a total of 8 predicates for the 3 types;
Table \ref{tab:statistics} lists these and gives statistics.

The dataset contains 65.3\% triple facts where the O entity is a multi-word phrase, and  58.6\% ambiguous facts where
the S or O entities share identical alias names in Wikidata.
For example, the two ambiguous entities ,\textit{``Birmingham, West Midlands (Q2256)''} and \textit{``Birmingham, Alabama (Q79867)''}, have the same \texttt{Label} value \textit{``Birmingham''}.
In total, 87.0\% of the sample facts have S entities in the long tail, where we define long-tail entities to have at most 13 Wikidata triples.

\section{Experimental Evaluation}

\vspace*{0.1cm}
\noindent
{\bf Baselines.}
To the best of our knowledge, there is no prior work on KBC or LM-as-KB that is specifically geared for coping with long-tail entities.
As a proxy, we thus compare to several state-of-the-art methods for relation extraction (RE) from text.
At test time, these methods receive the 
retrieved Wikipedia sentences for a ground-truth SPO fact and the SP pair as input, and are run to extract the withheld O argument (sentence-level extraction).

We compare to the following baselines:
\squishlist
\item \emph{NER + RC (CNN)} uses  TNER~\cite{ushio2022t} to recognize entity mentions in context sentences, followed by a CNN-based relation classifier ~\citet{nguyen2015relation}. The RC component is trained on  REBEL~\cite{cabot2021rebel}. 
\item \emph{REBEL}~\cite{cabot2021rebel} is an end-to-end relation extraction for more than 200 different relation types in Wikidata.
\item \emph{KnowGL}~\cite{knowgl-aaai_2023_demo} is an open-source system that can convert text into a set of Wikidata statements. 
\item \emph{GenIE}~\cite{josifoski-etal-2022-genie} is an end-to-end closed triplet extraction model, which is trained on REBEL dataset~\cite{cabot2021rebel}. GenIE uses Wikidata as the target KB and can extract 5,891,959 entities and 857 relations. 
\squishend

{\bf Setup.} 
There are two hyper-parameters for all competitors, the number of candidates $k$ (or the ``top-k'' hyper-parameter for baseline models)  and the threshold $\alpha$ for cutting off the extracted triples. 
For our framework, $k$ is  20 for all competitors and the threshold $\alpha$ is learned by using a hold-out (20\%) validation set. 
We report results for precision, recall and F1, with the original Wikidata triples as ground truth.
Although MALT provides canonicalized entities, we consider the extracted O to be a correct prediction as long as it appears in the alias table because some baselines themselves cannot do disambiguation.

Our method is completely unsupervised, and the only additional cost is prompt. We manually design one template for each relation (as shown in Table~\ref{tab:malt_template}).

\noindent
{\bf Results.}
Table \ref{tab:overall_performance} shows
the results from this experimental comparison.
We observe that the GenIE baselines does well in terms of precision, but has very poor recall.
In contrast, our two-stage method achieves both good precision and recall. Regarding precision, it is almost as good as GenIE (44\% vs. 52\%);
regarding recall, it outperforms GenIE and the other baselines by a large margin (43\% vs. 7\%).
Our method still leaves substantial room for further improvement, underlining the challenging nature of inferring facts for long-tail entities.
We think of our method as a building block to aid a human curator by judicious suggestions for facts that would augment the KG.

Many of the inferred SPO facts are indeed completely missing in Wikidata; so they are also not in the withheld ground-truth samples for the above evaluation. To estimate how many facts we could potentially add to the KG and how good our automatically inferred predictions are, we
picked 25 samples for each relation, a total of 250 fact candidates, and asked human annotators to 
assess their correctness. Over all relations, 
this achieved an average precision of 61\%.
For the relation {\small\tt educated at},
our method even has 76\% precision, and this is
a case where the KG has enormous gaps:
out of 10M sampled entities of type {\small\tt Human},
only 65\% have facts for this relation.
For this case, our KBC method collected 1.2M candidate facts,
showing the great potential towards closing these gaps.



\section{Conclusion}

We highlighted the challenge of knowledge base completion (KBC) for long-tail entities, introduced the MALT dataset for experimental comparisons and fostering further research, and presented a
completely unsupervised method for augmenting knowledge bases with long-tail facts.
Our method operates in two stages, candidate generation and candidate corroboration (incl. disambiguation), and leverages two different LMs in a complementary way.
Experimental results show substantial gains over state-of-the-art baselines, and highlight the benefits of our two-stage design with two LMs complementing each other.

\section*{Limitations}

Although our dataset presents a significant advancement over previous benchmarks, it is still limited in that it only contains entities already known to Wikidata. One could argue that the very long tail is what is even beyond Wikidata.

In the second stage, our method harnesses an LM pre-trained for entity disambiguation. Therefore, our methodology, in its current form, cannot predict objects that are not already known to that LM
and its underlying KB.

\section*{Acknowledgements}
This work was partially funded by ANR-20-CHIA-0012-01 (“NoRDF”). We thank Fabian M. Suchanek and Gaël Varoquaux for their helpful feedback.

\bibliography{acl2023}
\bibliographystyle{acl_natbib}

\appendix
\setcounter{table}{0}   
\setcounter{figure}{0}
\renewcommand{\thetable}{A\arabic{table}}
\renewcommand{\thefigure}{A\arabic{figure}}
\setcounter{equation}{0}
\section{Appendix}
\label{sec:appendix}
\subsection{The Motivation of Our Two-stage KBC Method}
In this section, we explain how we design the two-stage KBC method.
Existing approaches use cloze-style prompts to query masked language models. 
However, they cannot cope with multi-token facts well and suffer from the long-tail issue. 
Therefore, we experiment with a series of prompts for querying LMs, and experiments can be categorized into two classes: \emph{Context-Free} and \emph{Context-Based}. 

\begin{figure*}[t]
	\centering
	\includegraphics[width=1.0\textwidth]{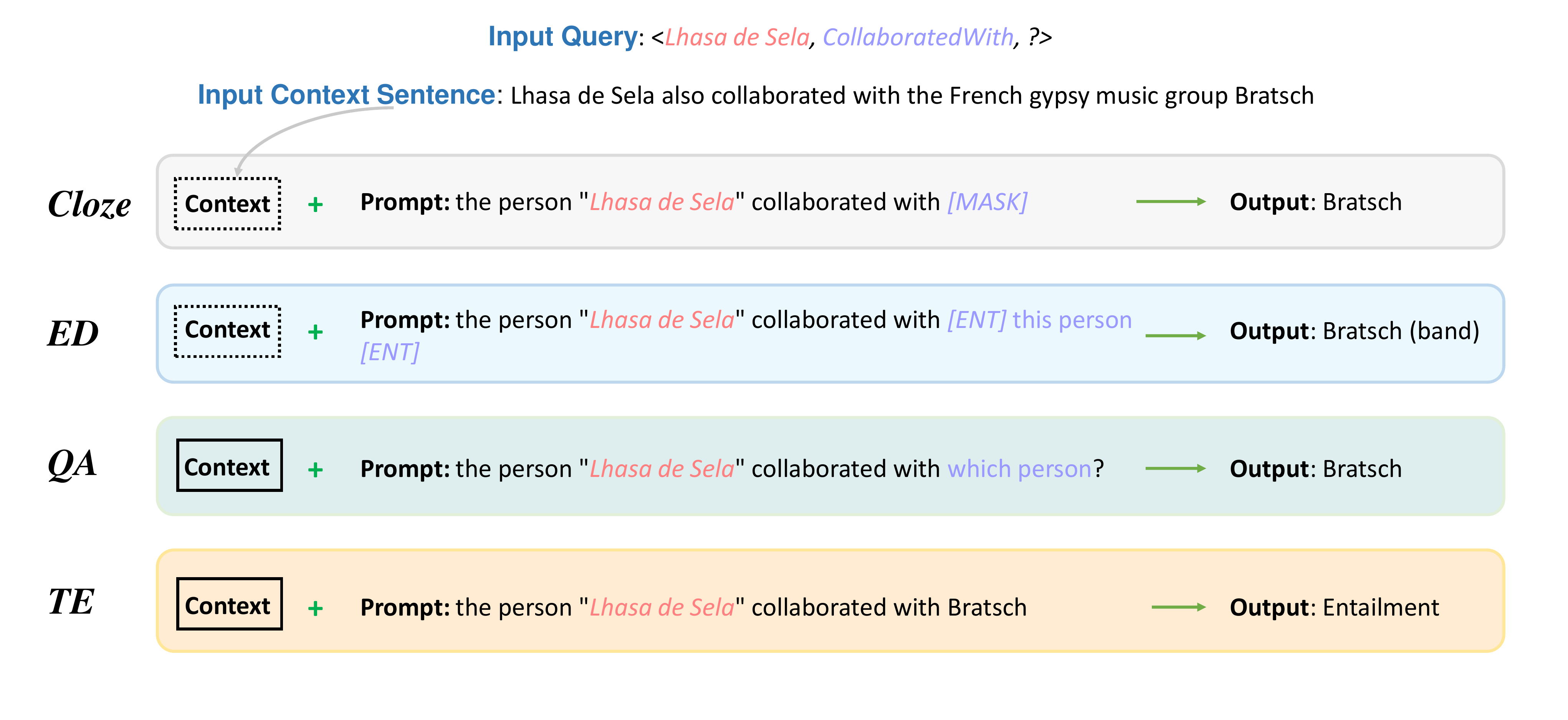}
	\caption{An illustration of different prompts for querying language models. The dashed lines mean the context sentence is optional. }
	\label{fig:diff_lms}
\end{figure*} 

Context-Free experiments evaluate the capabilities of LMs to generate facts by only using prompt queries. We
consider the following baselines. \\
\textbf{Cloze:} As prior methods, this baseline uses a cloze-style prompt to query masked LMs (the first frame in Figure~\ref{fig:diff_lms}). 
Here, two types of LMs are compared in this experiment. \textbf{Left-to-Right} LMs predict the upcoming words based on a sequence of words, and GPT-1~\cite{radford2018improving} and Transformer-xl~\cite{dai2019transformer} are used. 
\textbf{Masked} LMs aim to predict masked text pieces based on the surrounding context, and BERT-base and BERT-large~\cite{devlin2019bert} are used. 
To enable BERT to handle multi-token facts, we also introduce the decoding strategy proposed in X-FACTR~\cite{jiang2020x} for comparison.
\\
\textbf{ED:} Because the Cloze-style prompt cannot generate multi-token facts directly, we propose to use Language Models with Entity Disambiguators as knowledge bases, i.e., LMED-as-KB.
As shown in Figure~\ref{fig:diff_lms}, 
we can design such a prompt \textit{``the person Lhasa de Sela collaborated with \texttt{[ENT]} this person \texttt{[ENT]}.''}, where the mention is surrounded by special tokens \texttt{[ENT]} and \texttt{[ENT]}. 
After we use the prompt to query the generative disambiguation model, and it is able to disambiguate the mention \textit{``this person''} and output the correct canonicalized entity \textit{``Bratsch (band)''}, although the mention is not \textit{``this band''}. 
The core benefit of introducing LMED is that it can output disambiguated entity names with multiple tokens.
Here, we use the \textbf{Encoder-Decoder} entity disambiguation model GENRE~\cite{de2020autoregressive}, which is fine-tuned on BLINK~\cite{wu2020scalable} and AIDA~\cite{hoffart2011robust}.

In Context-based experiments, prompts are combined with additional context information to better retrieve facts from LMs, which has been demonstrated to substantially improve the cloze-style performance of LMs~\cite{petroni2020context}. 
Apart from Cloze and ED baselines, we introduce another two methods. \\
\textbf{QA:} Question-Answering models are able to extract answers to a question from a given document, and we adapt them to extract facts by designing question prompts. 
As shown in the third frame of Figure~\ref{fig:diff_lms}, given the input context and the question prompt \textit{``the person Lhasa de Sela collaborated with which person?''},  a QA model successfully outputs the correct answer. 
For experiments, we use two LMs fine-tuned  on the SQuAD 2.0~\cite{rajpurkar2018know}, RoBERTa-large~\cite{liu2019roberta}~\footnote{\url{https://huggingface.co/deepset/roberta-large-squad2}} and SpanBERT-large~\cite{joshi2020spanbert}~\footnote{\url{https://huggingface.co/mrm8488/spanbert-finetuned-squadv2}}. 
Besides, we use GPT3~\cite{brown2020language} as another QA baseline.
\\
\textbf{TE:} Textual Entailment models can judge whether a premise entails a hypothesis. 
To adapt TE for extracting facts from context, we first use a Named Entity Recognition model and then apply a textual entailment model to this entity and sentence for judging the entailment relation. For example,  given the context \textit{``Lhasa de Sela also appeared as a guest of the French gypsy music group Bratsch''}, the entity \textit{``Bratsch''} is recognized and we use the prompt: \textit{context $\rightarrow$ the person Lhasa de Sela collaborated with Bratsch}. If the premise entails the hypothesis, we can regard this as a correct tail entity. Here, we add type constraints for particular relations. Two LMs fine-tuned on TE datasets, RoBERTa-large~\cite{liu2019roberta}\footnote{\url{https://huggingface.co/ynie/roberta-large-snli_mnli_fever_anli_R1_R2_R3-nli}} and DeBERTa-large~\cite{he2020deberta}\footnote{\url{https://huggingface.co/MoritzLaurer/DeBERTa-v3-large-mnli-fever-anli-ling-wanli}},  are used in this experiment.
The NER model is TNER~\cite{ushio2022t}. 

Technically speaking, QA and TE are not LM-as-KB methods because they cannot generate facts without the help of context. 
However, these two methods have a unified pattern with Cloze and ED under the context-based setting, we, therefore, include them for comparison.

\begin{table}[tbp] 
   \tiny 
    \setlength{\tabcolsep}{1.2mm}{
	\begin{threeparttable} 
		\begin{tabular}{lccccccc}  
			\toprule    	
			Model&Prompt&Size&Multi-token&Disambiguated&P&R&F1\cr
			\midrule
                GPT-1&\textit{Cloze}&110M&\XSolidBrush&\XSolidBrush&0.3&3.2&0.7\cr
                Transformer-xl
                &\textit{Cloze}&257M&\XSolidBrush&\XSolidBrush&2.4&3.9&2.9\cr
                BERT- base&\textit{Cloze}&110M&\XSolidBrush&\XSolidBrush&7.1&4.9&4.2\cr
                  \quad \textit{w/ decoding}&\textit{Cloze}&110M&\Checkmark&\XSolidBrush&11.1&1.2&1.7\cr
                BERT-large&\textit{Cloze}&340M&\XSolidBrush&\XSolidBrush&19.0&3.7&4.7\cr
                \quad \textit{w/ decoding}&\textit{Cloze}&340M&\Checkmark&\XSolidBrush&8.7&2.1&2.4\cr
                \midrule
                GENRE&\textit{ED}&406M&\Checkmark&\Checkmark&19.1&5.4&7.4\cr
				\bottomrule
			\end{tabular}
                
			\caption{Context-Free performances of different language models on MALT.} 	\label{tab:only_prompt}
		\end{threeparttable}
	}
\end{table}

\begin{table}[tbp] 
   \tiny 
    \setlength{\tabcolsep}{1.1mm}{
	\begin{threeparttable} 
		\begin{tabular}{lcccccccc}  
			\toprule    	
			Model&Prompt&Size&Multi-token&Disambiguated&P&R&F1\cr
			\midrule
                BERT- base&\textit{Cloze}&110M&\XSolidBrush&\XSolidBrush&11.1&12.4&11.7\cr
                BERT- large&\textit{Cloze}&340M&\XSolidBrush&\XSolidBrush&11.8&14.4&12.3\cr

                \midrule
                RoBERTa-large&\textit{QA}&355M&\Checkmark&\XSolidBrush&5.6&45.1&9.7\cr
                 SpanBERT-large&\textit{QA}&340M&\Checkmark&\XSolidBrush&1.2&66.2&2.4\cr
                 GPT-3&\textit{QA}&175B&\Checkmark&\XSolidBrush&10.9&11.5&7.9\cr
                \midrule
                RoBERTa-large&\textit{TE}&355M&\Checkmark&\XSolidBrush&13.2&19.7&13.4\cr
                DeBERTa-large&\textit{TE}&304M&\Checkmark&\XSolidBrush&13.5&22.2&14.6\cr
                \midrule
                GENRE&\textit{ED}&406M&\Checkmark&\Checkmark&16.5&30.9&18.9\cr
                \bottomrule
			\end{tabular}
                
			\caption{Context-Based performances of different language models on MALT.} 	\label{tab:with_context}
		\end{threeparttable}
	}
\end{table}

\subsubsection{Can LMs Generate Facts?}
In this context-free experiment, we aim to answer whether LMs can generate facts and various models are evaluated on MALT.
The experimental results are shown in Table~\ref{tab:only_prompt}. 
We first observe all models perform poorly on MALT-Wikidata because it contains a large number of multi-token and long-tail entities. 
Left-to-Right and Masked LMs have difficulties in dealing with these facts, even with the introduction of multi-token decoding.
Moreover, we observe that GENRE outperforms other baselines consistently and this confirms the feasibility of the usage of LMED-as-LM. 
Overall, a single query does not retrieve facts from LMs very effectively, and the reasons are twofold: 1) the capacity of LMs
for storing world knowledge is limited by model size, i.e., LMs with tens or hundreds of billions of parameters can memorize all facts in Wikidata~\cite{heinzerling2021language}; 
2) proper prompts are needed for a better recall, e.g., by additional information or prompt engineering.

\begin{figure}[t]
	\centering
	\includegraphics[width=0.5\textwidth]{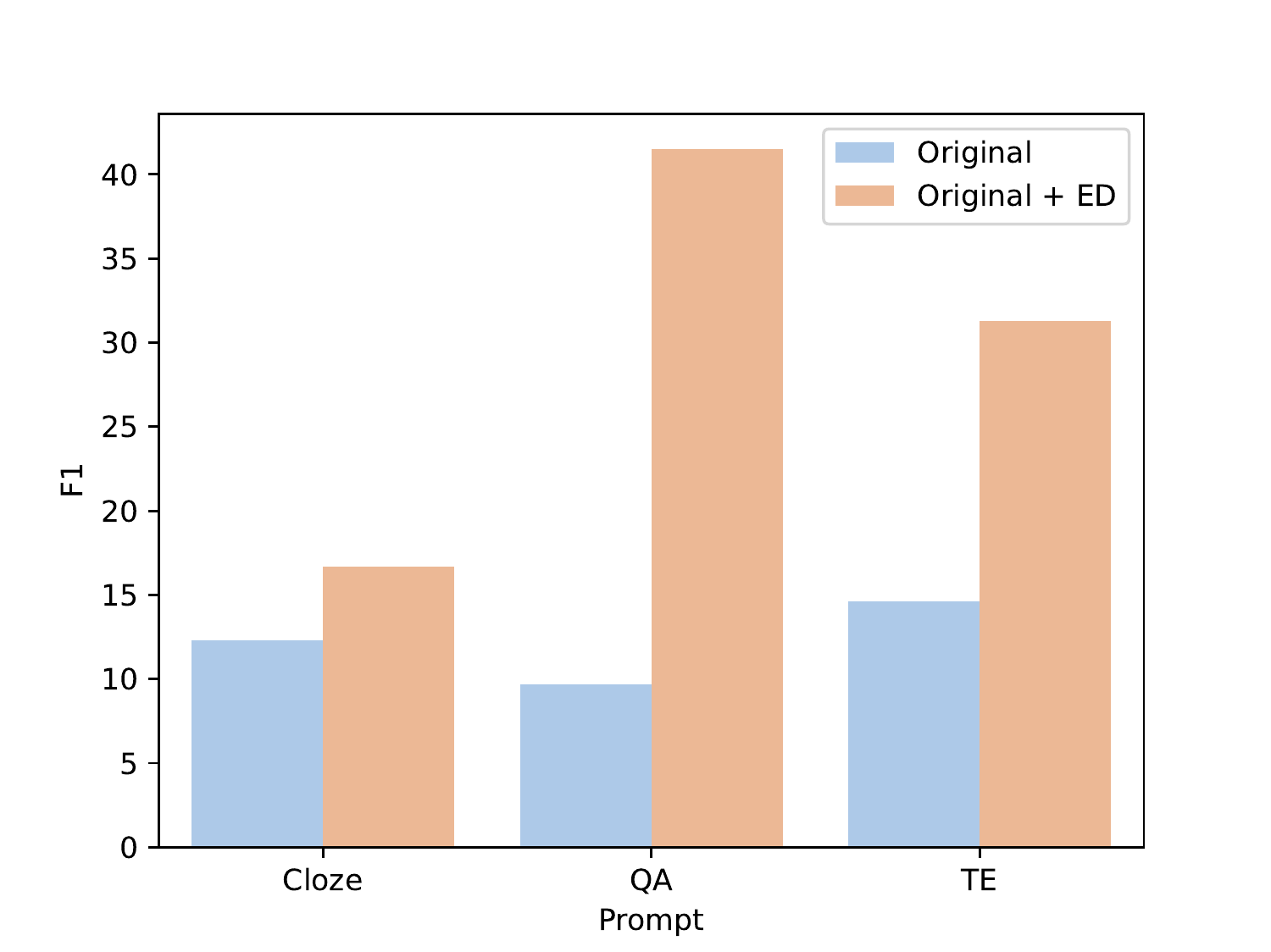}
	\caption{Improvements of adding the ED Prompt. }
	\label{fig:baseline_with_ed}
\end{figure}

\subsubsection{Can Context help?}
In this context-based experiment, context sentences are introduced for assessing the capability of LMs to
generate facts by exploiting context.
Concretely, we traverse the sentences in Wikipedia for relevant entities and each context sentence is combined with a corresponding prompt to compose a new query. 
Next, facts are retrieved or extracted by using different LMs.
For duplicated outputs, we merge them and average the score.
The experimental results are shown in Table~\ref{tab:with_context}.
We can see that adding context can remarkably improve the performances on MALT-Wikidata, e.g., BERT-large (4.7 $\rightarrow$ 12.3) and GENRE (7.4 $\rightarrow$ 18.9 ). 
GENRE consistently outperforms other baselines in terms of F1 while QA mode can obtain very high recalls. 
For TE methods,  they are a workable approach while still lagging behind our framework.

\subsubsection{Our Two-stage KBC Method}
Based on the above analyses, we find that ED prompts can generate disambiguated and relatively high-quality  facts while QA prompts have the highest recall. 
Hence, a question naturally appears: \textit{``Can we synergize the two components to yield better facts?''}

To answer this question, we apply the ED prompt method to the facts generated by the other three methods, Cloze, QA, and TE. 
The post-processing step of ED prompt serves to verify and re-rank the candidates of the first step. 
The experimental results are shown in Figure~\ref{fig:baseline_with_ed}. We observe the combination can bring consistent improvements and the pipeline of ``QA + ED'' achieves the best score. 
Therefore, we leverage two different LMs in a two-stage pipeline.
The first stage generates candidate answers by using a high-recall question-answering model.
The second stage employs an entity disambiguation model for validating the candidates.

\end{document}